\documentclass[letterpaper, 10 pt, conference]{ICRA/ieeeconf}  

\IEEEoverridecommandlockouts                              

\usepackage{graphics} 
\usepackage{epsfig} 
\usepackage{mathptmx} 
\usepackage{times} 
\usepackage{amsmath} 
\usepackage{amssymb}  
\usepackage{float}

\usepackage{tikz}
\usepackage{todonotes}
\usetikzlibrary{shapes.geometric, arrows.meta, positioning}

\title{\LARGE \bf
From Bench to Flight: Translating Drone Impact Tests into Operational Safety Limits
}

\author{Aziz Mohamed Mili$^{1}$, Louis Catar$^{1}$, 
Paul Gérard$^{1}$,  Ilyass Tabiai$^{1}$ and David St-Onge$^{1}$
\thanks{*This work was supported by the NSERC CREATE UTILI program and FRQNT Team grant (\#283381)}
\thanks{$^{1}$All authors are with Lab INIT Robots, Department of Mechanical Engineering, 
        École de technologie supérieure, Montreal, QC, Canada
        {\tt\small david.st-onge@etsmtl.ca}}%
}

\begin{document}

\maketitle
\thispagestyle{empty}
\pagestyle{empty}

\begin{abstract}
Indoor micro-aerial vehicles (MAVs) are increasingly considered for tasks that bring them in close proximity to people, yet practitioners lack a practical way to tune motion limits to measured impact risk. We present an end-to-end, open toolchain that turns benchtop impact tests into deployable safety governors for drones. First, we detail a compact, replicable impact rig and protocol that captures force–time profiles across drone classes and contact surfaces. Second, we provide data-driven fits that map pre-impact speed to impulse and contact duration, enabling direct computation of speed bounds for a target force limit. Third, we release scripts and a ROS2 node that enforce these bounds online and log compliance, with hooks for facility-specific policies. We validate the workflow on multiple commercial off-the-shelf quadrotors and representative indoor assets, demonstrating that the derived governors preserve task throughput while meeting force constraints specified by safety stakeholders. The contribution is a practical bridge: from measured impacts to runtime limits, with sharable datasets, code, and a repeatable process that teams can adopt to certify indoor MAV operations near humans.
\end{abstract}

\section{INTRODUCTION}

Uncrewed aerial systems (UAS) are increasingly deployed in indoor environments for tasks ranging from warehouse inventory management~\cite{proia2022} to infrastructure inspection ~\cite{dong2024design} and emergency response in confined spaces. These scenarios inherently require close human–robot proximity, where conventional ``keep-out'' safety paradigms designed for outdoor aviation become less practical. As a result, collision safety at operational speeds of 2–10 m/s has emerged as a key concern for certifying micro-aerial vehicle (MAV) operations in human-populated environments. Regulatory and standards efforts reflect this need, including JARUS AMC RPAS.1309~\cite{jarus2019}, Transport Canada AC 922-001~\cite{tc922001}, and EASA’s Specific Category/SORA framework~\cite{easa2019}.
 mandate operation-centric risk assessment and evidence that mitigations such as speed/energy limits, geofencing, containment, and protective design meet safety objectives—but they leave manufacturers to supply the underlying data and models. ISO/TS~15066 collaborative force thresholds~\cite{iso15066} provide general safety principles in human-robot collaborative tasks, yet they were not developed with MAV-specific impact conditions in mind. Meanwhile, researchers and companies tackle the impact-resilient and human-safe drone design challenge without a clear set of requirements and benchmark solutions.

This paper addresses the gap between experimental insight and deployable safety systems. We contribute what we believe is the first system-level, rebound-aware, indoor-speed impact dataset on complete drones, quantify how orientation and structural compliance reshape the peak-force–rebound trade-off at representative speeds, and translate these empirical regressions into a ROS~2 safety governor that limits velocity to satisfy human-contact force bounds—backed by measurements, not finite-element models. Rather than replacing existing frameworks, we provide a practical toolchain that connects impact physics to runtime control, enabling certifiable, facility-specific speed governance for indoor MAV operations near people. The remainder of the paper presents the benchtop apparatus and protocol and the derived observables, details the regression models and the governor design, and validates the approach across multiple platforms and contact orientations before discussing limitations, safety implications, and artifact release.

\section{RELATED WORK}

Research on MAV safety has advanced along three axes: impact characterization, safety standards, and protective structures. While each has progressed significantly, they remain largely unintegrated.


\begin{figure*}[ht]
    \centering
    \includegraphics[width=0.9\textwidth]{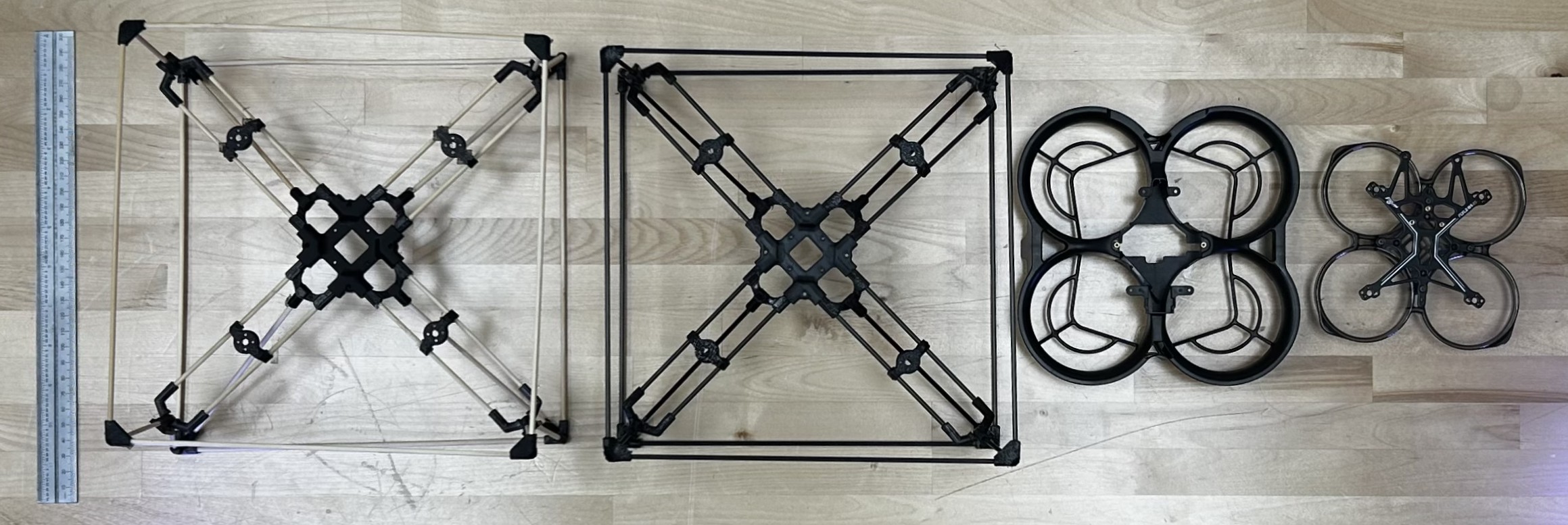}
    \caption{UAVs considered for this study, from left to right: Bamboo Cognifly, Carbon Cognifly, DJI Avata, and Flywoo Flylens.}
    \label{fig:placeholder}
\end{figure*}

\noindent\textbf{Impact Characterization} The earliest attempts to assess drone collisions adapted methodologies from manned aviation. The FAA-funded ASSURE program~\cite{assure2019} produced the most comprehensive dataset to date, employing anthropomorphic test devices (ATDs) and post-mortem human subjects (PMHS) to generate injury probability curves. However, their reliance on automotive metrics such as Head Injury Criterion (HIC) and Abbreviated Injury Scale (AIS), tuned for impacts above 10~m/s, limits direct applicability to indoor MAV lower speeds. Berthe et al.~\cite{berthe2019} further exposed inconsistencies when applying such scales at drone speeds of 16~m/s.

More recent efforts refined the metrics. Svatý et al.~\cite{svaty2022} showed that blunt trauma criteria systematically overestimate risk for MAVs below 25~kg. Bai et al.~\cite{bai2025} introduced a four-grade safety classification based on drop tests across drones ranging from 0.43 to 4.53~kg, yet vertical drops omit the oblique approaches typical of indoor collisions. Catar et al.~\cite{catar2025} advanced methodology by catapult-launching ultra-light polymer micro-lattices, achieving specific energy absorption above 1000~J/kg and highlighting rebound as a neglected safety parameter. These studies collectively indicate that MAV safety cannot be reduced to energy thresholds alone but requires time-resolved impact characterization.

\noindent\textbf{Standards and Regulatory Frameworks} In parallel, standards bodies have moved cautiously toward formalization. EASA’s Specific Category regulations~\cite{easa2019} require collision risk mitigation, while IEEE has issued UAS-related standards on quality of service and payload interfaces~\cite{ieee_standards}. ISO/TS~15066~\cite{iso15066}, though designed for collaborative robots, provides explicit force thresholds directly relevant to human–MAV contact, such as for the face (65N), neck (150N), chest (140N), back and shoulders (210N). This technical specification also provides guidelines on zone task definition around the user. Proia et al.~\cite{proia2022} demonstrated how such limits could guide safe drone operation in warehouses. Still, neither ISO nor IEEE offer validated test protocols tailored to MAV impacts, leaving a gap between compliance requirements and reproducible laboratory data.



\noindent\textbf{Protective Structures} Structural innovation has become a parallel thread. Expandable cages such as PufferBot~\cite{hedayati2020} create deployable safety buffers. Park et al.~\cite{park2023} introduced origami–kirigami guards actuated by shape memory alloys, reducing side-impact force by 78\%. Quek et al.~\cite{quek2022} designed foldable winged shells that autorotate during free fall, while Aloui et al.~\cite{aloui2025} demonstrated tensegrity-based drones where elastic lattices dissipate impact energy without structural failure. At the material level, Catar et al.~\cite{catar2025}’s micro-lattices achieve record energy absorption at densities as low as 65~kg/m$^3$. Complementary to these, Patnaik et al.~\cite{patnaik2024} developed XPLORER, a passive deformable quadrotor that leverages chassis compliance for agile, contact-rich navigation. These works reveal a vibrant design space, but evaluations are usually structural-only and not integrated with safety-aware control.

\section{METHODS AND MATERIAL}
\subsection{Custom Impact bench}
\begin{figure*}[!ht]
\begin {center}
\includegraphics[width=0.75\textwidth]{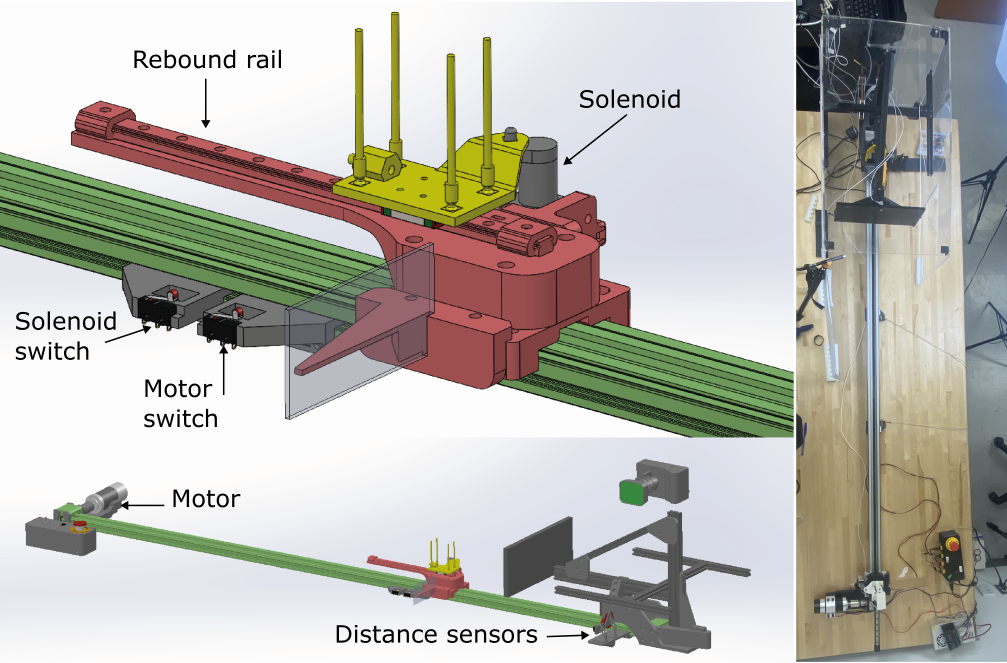}
\caption{Overview of the custom impact test bench. The main rail (green) is fixed and used to propel the mobile rail (red). The sample trolley (yellow) holds the drone. Just before impact, it is released by the solenoid (gray) and translates freely along the red rail. }
\label{fig:bench}
\end {center}
\end{figure*}

We designed a custom impact test bench for this study, shown in Fig.~\ref{fig:bench}. It consists of a linear catapult driven by an electric motor, enabling the testing of aerial vehicle structures or complete micro-UAS under conditions close to their real-world use. The bench facilitates controlled-speed impacts. At launch, a propulsion trolley is accelerated along the main rail. Just before impact, a drone-holding trolley, initially fixed to the propulsion trolley, is released onto its secondary rail (located atop the propulsion trolley). This design allows the drone to rebound freely upon impact. 

The current study uses full drone structures directly mounted on a custom ultra-light support. This support, made of carbon fiber and rigid materials, is specifically designed to minimize its influence on the overall inertia of the tested system as well as the drone's dynamic behavior during collisions. All components are optimized to keep the mass as low as possible, granting more freedom to replicate real flight conditions (for instance adding calibrated weights).

Impacts are conducted at a speed between \textbf{3 - 4m/s}: a compromise between high cruising velocity and slow scene scanning velocity. The current rebound zone, limited to 300 mm, do not allow proper measurement of larger rebounds at higher speeds. This speed selection maximizes impact velocity while maintaining an exploitable rebound zone within current hardware limitations, while still representing realistic indoor navigation scenarios.

To characterize the impact events, we employed a suite of sensors integrated into the test bench. Forces were recorded using three PCB Piezotronics\textsuperscript{\copyright} load cells positioned in a triangular configuration behind the impact wall. A PCB Piezotronics\textsuperscript{\copyright} single-axis accelerometer was mounted on the mobile trolley to capture deceleration, and a TFmini-S operating at 1000Hz measured velocity before impact. Each collision was also filmed with a Chronos 2.0 high-speed camera at 1000 frames per second, with a background grid to aid in analyzing deformation patterns and rebound speed.

Analog signals from the force and acceleration sensors were fed into a Siemens\textsuperscript{\copyright} SCADAS acquisition system, sampled at 6250~Hz. The distance sensors, meanwhile, were read via a Python script on a Linux laptop. Both systems were synchronized using a 5V trigger pulse sent just before launch, allowing precise temporal alignment of all data streams.



\subsection{Test Specimens}
This experimental campaign examines how small drones behave under low-speed impact conditions, focusing on energy absorption and rebound. The core hypothesis is that safer drones, both for humans and the machines themselves, are those that either absorb more kinetic energy or rebound more effectively upon collision.

Twenty drone specimens were tested, combining both commercial models and custom-built structures. Among the commercial units, we included 4 DJI Avata drones, known for their integrated propeller guards and moderate cruising speed of 8~m/s, and 4 Flywoo FlyLens 85 Whoop drones, which are faster and lighter, reaching up to 20~m/s. The custom-built models, Cognifly~\cite{azambuja2022cognifly}, were designed in-house and fabricated using either carbon rods or bamboo rods, both combined with flexible TPU joints. We tested 8 carbon-based Cognifly units and 4 in bamboo, ensuring uniform construction for consistency. Figure~\ref{fig:placeholder} shows the four drone configurations used in this study.

Each drone was mounted on a lightweight mobile platform using elastic fasteners. 

\subsection{Material Characterization}

Understanding the material composition and mechanical properties of each drone configuration is essential for interpreting impact behavior and energy absorption mechanisms. The test specimens represent distinct material approaches to lightweight aerial vehicle design, each with different strategies for balancing structural integrity and impact safety.

Table~\ref{tab:material_properties} summarizes the key mechanical properties and design characteristics of each configuration.


\begin{table*}[h]
\centering
\caption{Material Properties and Design Characteristics of Test Specimens}
\label{tab:material_properties}
\begin{tabular}{|c||c|c|c|c|c|c|}
\hline
\rule{0pt}{3ex}\textbf{Configuration} & \textbf{Frame Material} & \textbf{Mass} & \textbf{Young's Modulus} & \textbf{Tensile Strength} & \textbf{Density} & \textbf{Toughness}\rule[-1ex]{0pt}{0pt} \\
& & \textbf{(g)} & \textbf{(GPa)} & \textbf{(MPa)} & \textbf{(g/cm$^3$)} & \textbf{(kJ/m$^2$)} \\
\hline
\rule{0pt}{2.5ex}DJI Avata & Nylon + Glass fiber & 410 & 8--12 & 120--180 & 1.35--1.40 & 80--120\rule[-1ex]{0pt}{0pt} \\
& (PA66-GF30) & & & & & \\
\hline
\rule{0pt}{2.5ex}Flywoo FlyLens & Carbon fiber CFRP & 250 & 230 & 1600--2300 & 1.4--1.8 & 40--60\rule[-1ex]{0pt}{0pt} \\
& (Rigid joints) & & & & & \\
\hline
\rule{0pt}{2.5ex}Cognifly Carbon & Carbon fiber rods & 270 & 230 & 1600--2300 & 1.4--1.8 & 40--60\rule[-1ex]{0pt}{0pt} \\
& (TPU joints) & & & & & \\
\hline
\rule{0pt}{2.5ex}Cognifly Bamboo & Natural bamboo fiber & 250 & 15--35 & 400--800 & 1.35--1.38 & 15--30\rule[-1ex]{0pt}{0pt} \\
& (TPU joints) & & & & & \\
\hline
\end{tabular}
\vspace{-2em}
\end{table*}

The material property ranges reflect typical values for these composite systems under quasi-static loading conditions. Dynamic loading effects, particularly relevant for impact scenarios, generally increase stiffness and strength for polymer-matrix composites due to viscoelastic behavior, while natural fiber composites like bamboo exhibit more complex rate-dependent responses due to their cellular microstructure.

\subsection{Impact Conditions and calibration}

All impacts were performed at a speed ranging from 3~ to 4~m/s on our custom-designed impact bench with rebound dynamics captured.

Most tests were conducted under frontal impact conditions, with the drone aligned perpendicularly to the wall and parallel to the ground. For the carbon Cognifly drones, an additional configuration was tested: a 45-degree vertical tilt, designed to simulate oblique impacts and explore how angle affects energy dissipation and post-impact motion.

Raw sensor data were processed using a multi-stage validation and filtering approach to extract reliable impact parameters while maintaining conservative safety estimates for robotic applications.

\textbf{Stage 1: Synchronization and Validation}
Multi-sensor data acquisition employed hardware-triggered simultaneous recording (5V TTL pulse) ensuring temporal alignment within $\pm0.1$~ms. 

\textbf{Stage 2: Noise Reduction and State Estimation}
Velocity estimation employed dual-stage filtering:
\begin{enumerate}
\item \textbf{Median filtering}: 5-point sliding window with outlier detection based on $3\sigma$ deviations from local median
\item \textbf{Kalman filtering}: State-space estimation with state vector $\mathbf{x} = [p, v]^T$
\end{enumerate}

The Kalman filter used conservative discrete process noise $Q$ for a constant acceleration model $F_{2\times2}$ with system variance $\sigma_s=0.01$.

Force signals underwent 4th-order Butterworth low-pass filtering (1000~Hz cutoff) to remove high-frequency noise while preserving impact dynamics.

\textbf{Stage 3: Impact Detection}
Impact detection combined force thresholds ($F_{\text{norm}} > 6$~N) with velocity validation ($|v| > 2.5$~m/s) and high-speed camera confirmation to prevent false positive detections.

\textbf{Stage 4: Conservative Force Processing}
During impact, elastic coupling between drone structures and test bench creates measurement artifacts including negative force readings during continued contact. To address these while maintaining conservative estimates, impulse integration employed positive-force rectification:

\begin{equation}
J = \int_{t_{\text{impact}}}^{t_{\text{end}}} F^+(t) \, dt
\end{equation}

where $F^+(t) = \max(F(t), 0)$. This approach provides lower-bound impulse estimates suitable for safety applications by excluding potentially non-physical negative contributions.

\begin{figure}[h]
    \centering
    \includegraphics[width=\linewidth]{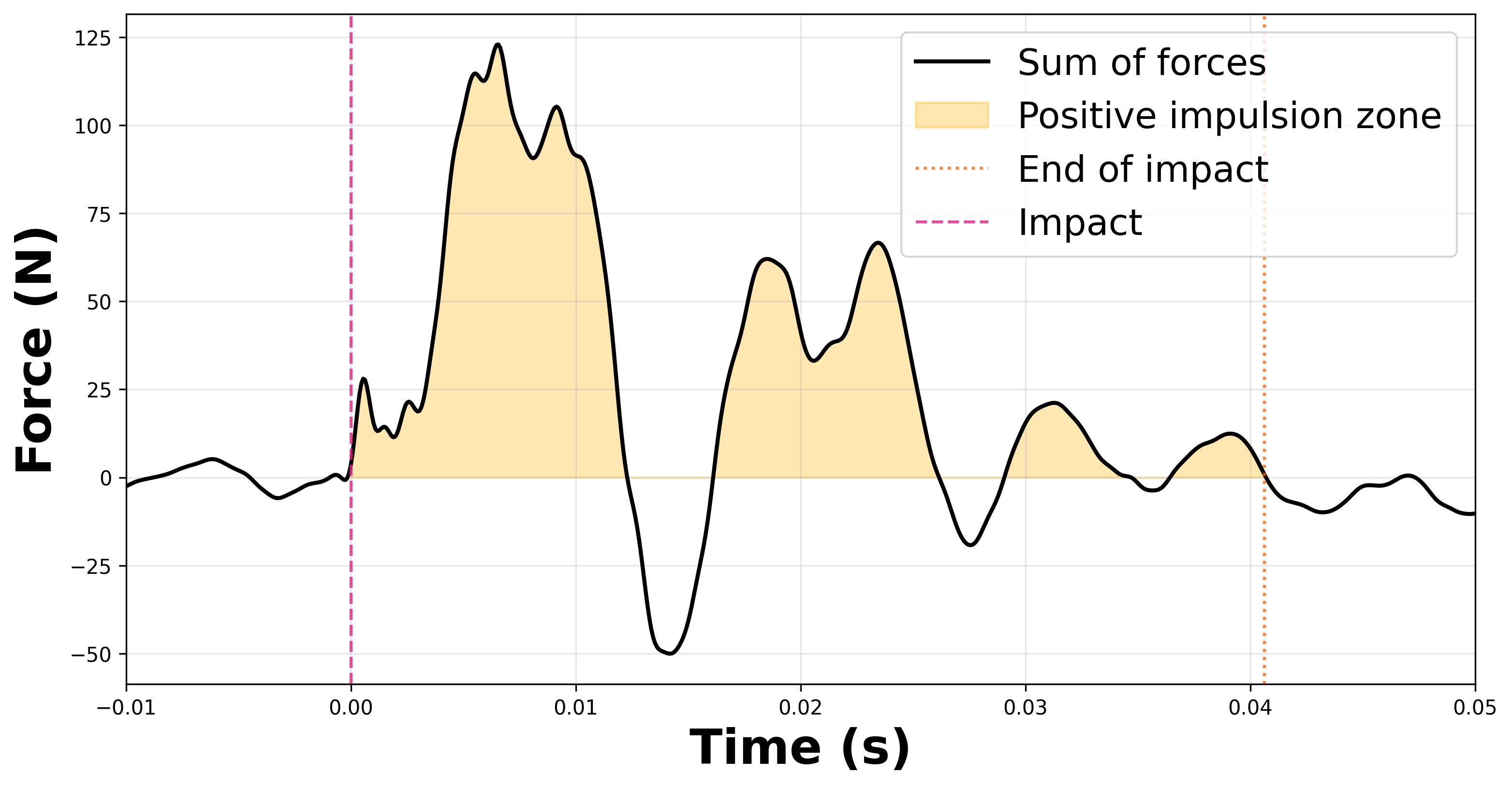}
    \caption{Force data sample from a carbon Cognifly impact showing at least 3 stages impact with restitution of the force plate in-between.}
    \label{fig:placeholder}
\end{figure}

Impact duration was determined via high-speed video analysis and compared to force zero-crossings to ensure timing accuracy independent of measurement artifacts.

The processing pipeline prioritized safety over precision through: conservative Kalman parameters preventing over-smoothing, strict outlier detection maintaining data integrity, multi-modal impact validation preventing false negatives, and positive-force integration providing safety margins. This ensures the extracted parameters represent conservative estimates suitable for robotic safety system integration.

\subsection{Statistical Analysis and Curve Fitting}

To deploy the results from our impact campaign in practical settings, we evaluate velocity-dependent impact relationships using polynomial regression. Given the limited sample size ($n=4$ per configuration), we employed dual-metric validation: 1. the coefficient of determination ($R^2$) and 2. Mean Absolute Error (MAE) for robust curve fitting assessment.

Traditional $R^2$ values can be misleadingly low in small-sample experimental datasets despite accurate physical representation. MAE provides absolute error quantification essential for safety governor applications requiring precise impact prediction at various velocities.

\subsection{Safety Governor force estimation}

\begin{figure}[h]
    \centering
    \includegraphics[width=\linewidth]{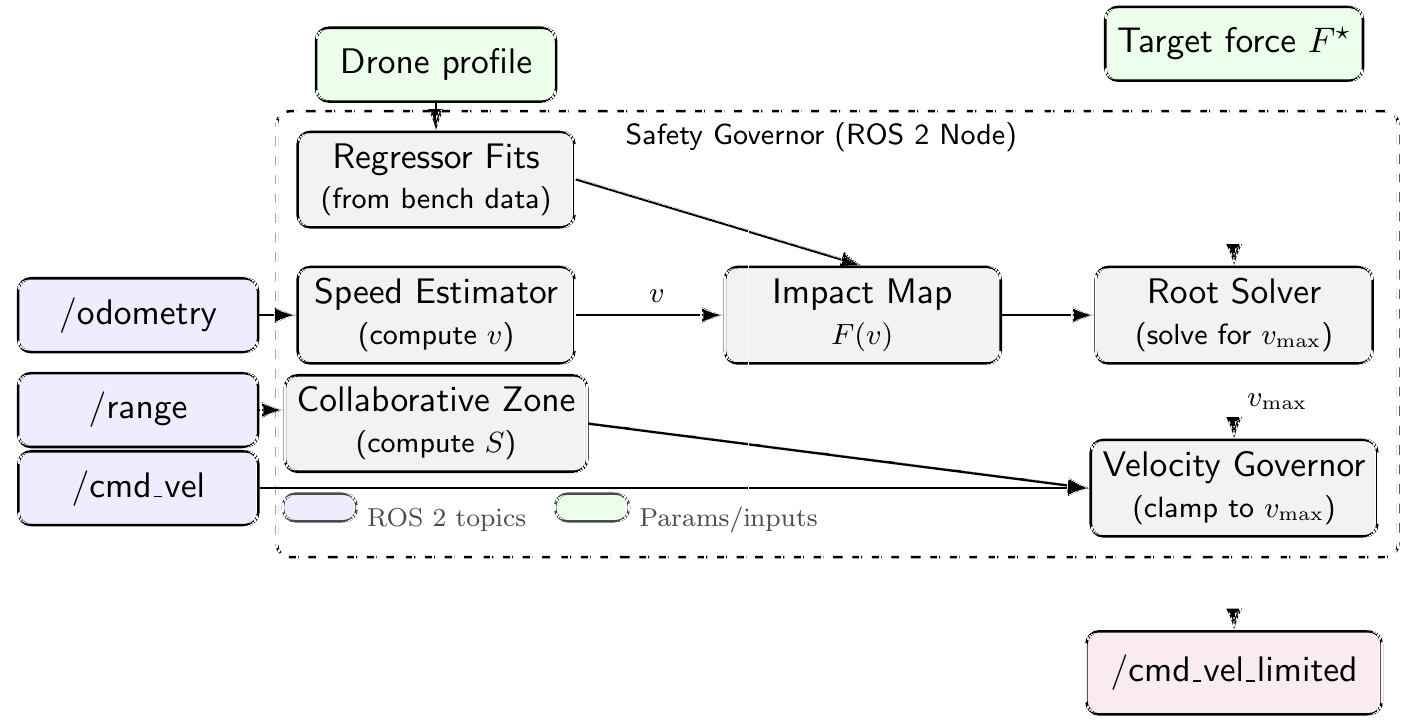}
    \caption{Safety governor block diagram: \texttt{/odometry} and \texttt{/range} yield speed $v$; bench-derived regressors define the impact map $F(v)$; a root-solver computes $v_{\max}$ for a target force; the governor publishes \texttt{/cmd\_vel\_limited}.}
    \label{fig:sg}
\end{figure}

Vision-based detection frameworks~\cite{aguilar2017,padhy2019} enable human-aware path planners, but remain vulnerable to occlusions, lighting changes, external disturbances, and fast, unpredictable human motion. To complement these methods, we introduce a Safety Governor: a lightweight ROS 2 layer that enforces conservative speed limits whenever people are nearby, independent of the upstream planner. The governor derives its limits from the measured impact behavior of the specific airframe, converting bench-fit regressions into runtime velocity caps that satisfy a target contact-force bound. By saturating \texttt{/cmd\_vel} to \texttt{/cmd\_vel\_limited} (illustrated in Fig.~\ref{fig:sg}), it provides a last-line safeguard that maintains safety even when perception or tracking briefly degrades.

Similar to Proia et al.~\cite{proia2022} ISO/TS~15066 integration, we first compute the radius of collaboration zone around the human as
\begin{equation}
S(t_0) \;=\; \underbrace{v\,T_q}_{\text{perception \& response}} \;+\; \underbrace{v\,T_s}_{\text{braking onset}} \;+\; \underbrace{B}_{\text{braking distance}} \;+\; C,
\label{eq:iso-radius}
\end{equation}
where $v$ is the current UAV speed within the collaborative zone, $T_q$ is the end-to-end response latency
(perception, planning, and actuation), $T_s$ is the time to stop, $B$ the distance traveled during active braking,
and $C$ a margin capturing expected human reach. With the worst-case deceleration $a>0$,
$T_s = v/a$ and $B = v^2/(2a)$, so \eqref{eq:iso-radius} reduces to
\begin{equation}
S(t_0) \;=\; v\,T_q \;+\; \frac{3}{2}\,\frac{v^2}{a} \;+\; C.
\label{eq:iso-compact}
\end{equation}

When in the collaboration zone, we expect the human to move in unpredictable and fast manners. To ensure safe mission we here restrict the UAV dynamics below a maximum impact force. We bound speed by a force-based cap derived from benchtop impact data.
Let $m$ be the mass, $v>0$ the approach speed, $E_r(v)=\bigl(v_f/v\bigr)^2$ the kinetic-energy restitution,
and $\hat e(v)=\sqrt{E_r(v)}$; with the measured contact-duration regression $\widehat{\Delta t}(v)$, the
average impact force is
\begin{equation}
F_{\mathrm{avg}}(v) \;=\; \frac{m\,v\bigl(1+\hat e(v)\bigr)}{\widehat{\Delta t}(v)}.
\label{eq:Favg}
\end{equation}
Given a target force limit $F^\star$ (average or peak mapped to average), we solve the scalar root
\begin{equation}
\frac{m\,v\bigl(1+\hat e(v)\bigr)}{\widehat{\Delta t}(v)} \;-\; F^\star \;=\; 0
\label{eq:force-root}
\end{equation}
to obtain the force-constrained speed $v$.

At runtime, the ROS~2 governor fuses the distance-aware ISO cap with the impact-based cap, publishing the saturated command on \texttt{/cmd\_vel\_limited}. This construction preserves compliance with
iso-style collaborative radii while guaranteeing that any residual contact obeys the empirically validated
force limit for the specific airframe.


\section{RESULTS}

This section presents the experimental characterization of drone impact dynamics across multiple models, and their direct implications for safety governor design in human-proximate operations. Results are presented as mean values and standard deviations from four independent trials per configuration (n=4).

\begin{table*}[h]
\caption{Experimental Results of Drone Impact Tests -- Mean Values and Standard Deviations}
\label{tab:impact_results}
\begin{center}
\begin{tabular}{|c||c|c|c|c|c|}
\hline
\rule{0pt}{3ex}\textbf{Configuration} & $F_{max}$ (N) & $\Delta t_J$ (ms) & $J$ (N$\cdot$s) & $EC_i$ (J) & $EC_r$ (\%)\rule[-1ex]{0pt}{0pt} \\ 
\hline
\rule{0pt}{2.5ex}Bamboo-0°    & 84.4 $\pm$ 3.3   & 41.4 $\pm$ 11.3 & 1.181 $\pm$ 0.222 & 1.42 $\pm$ 0.09 & 16.4 $\pm$ 2.9\rule[-1ex]{0pt}{0pt} \\
\hline
\rule{0pt}{2.5ex}Carbon-0°    & 105.6 $\pm$ 7.0    & 36.0 $\pm$ 3.3 & 1.144 $\pm$ 0.081 & 1.42 $\pm$ 0.06 & 14.6 $\pm$ 0.5\rule[-1ex]{0pt}{0pt} \\
\hline
\rule{0pt}{2.5ex}Flywoo-0°    & 120.8 $\pm$ 11.7   & 17.2 $\pm$ 0.2 & 1.192 $\pm$ 0.057 & 1.57 $\pm$ 0.01 & 12.8 $\pm$ 0.4\rule[-1ex]{0pt}{0pt} \\
\hline
\rule{0pt}{2.5ex}Avata-0°     & 230.4 $\pm$ 27.3 & 22.1 $\pm$ 0.9 & 1.846 $\pm$ 0.311 & 2.41 $\pm$ 0.15 & 14.9 $\pm$ 0.9\rule[-1ex]{0pt}{0pt} \\
\hline
\hline
\rule{0pt}{2.5ex}Carbon-45°   & 133.9 $\pm$ 30.0  & 38.3 $\pm$ 1.9 & 1.468 $\pm$ 0.204 & 1.16 $\pm$ 0.09 & 11.2 $\pm$ 0.1\rule[-1ex]{0pt}{0pt} \\
\hline
\end{tabular}
\end{center}
\vspace{-2em}
\end{table*}

\subsection{Baseline Frontal Impact Performance (0° Configuration)}

\subsubsection{Peak Impact Force Distribution and Safety Thresholds}

Peak impact forces follow a clear hierarchy strongly correlated with structural mass and compliance mechanisms (Table~\ref{tab:impact_results}). The bamboo Cognifly configuration achieves the lowest peak force (84.4 $\pm$ 3.3 N). While this value lies below the ISO/TS 15066 threshold for the neck (150 N), but still exceeds the more restrictive limit for the face (65 N), meaning that none of the tested configurations can be considered fully safe for facial impacts. By contrast, the DJI Avata generates 230.4 $\pm$ 27.3 N, surpassing not only the face and neck limits but also those of the chest (140 N), though remaining below the back/shoulder threshold of 210 N~\cite{iso15066}.

\begin{figure}[h]
    \centering
    \includegraphics[width=\linewidth]{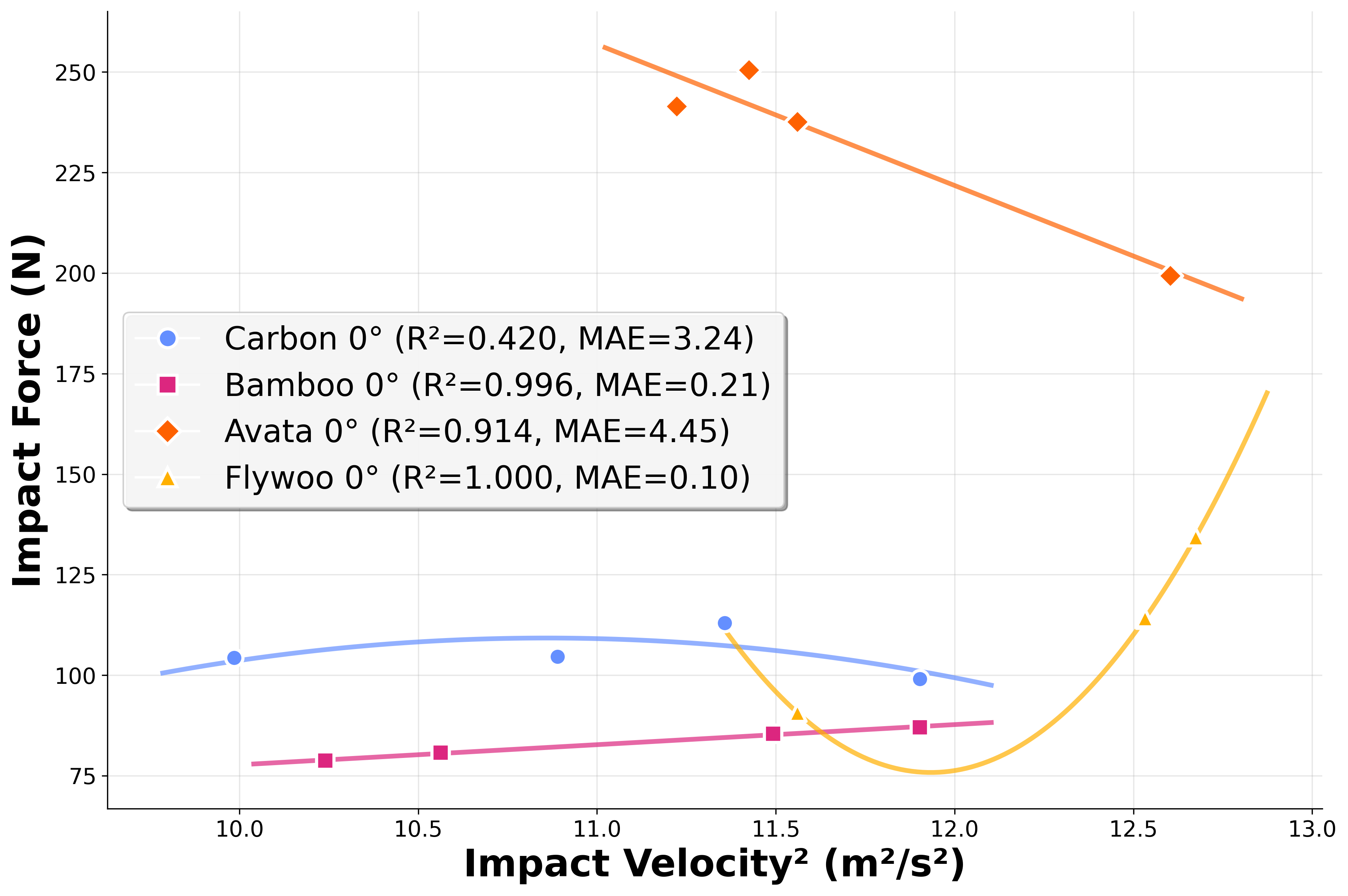}
    \caption{Peak impact forces at 0° orientation}
    \label{fig:force_0deg}
\end{figure}

For intermediate designs, compliance proves decisive. The carbon Cognifly, despite using the same carbon fiber as the rigid Flywoo, achieves a 25\% reduction in peak force (105.6 vs 140.8 N) thanks to its TPU joints, placing it within the safe range for the chest and shoulders but not for the face. These differences are statistically significant and highlight that structural compliance can shift MAV impacts from clearly unsafe to conditionally acceptable, depending on the body region affected. Such insights are critical for informing safety governor thresholds and for aligning drone design with human-robot interaction standards.

\subsubsection{Contact Duration Independence from Velocity}

The experimental results in Table~\ref{tab:impact_results} show that contact duration remains largely independent of impact velocity across all tested configurations. Rigid designs such as Flywoo exhibit very stable durations (17.2 $\pm$ 0.2 ms), while more compliant structures like the bamboo-based Cogni show higher variability (41.4 $\pm$ 11.3 ms) due to their nonlinear deformation behavior and discontinuous bending.  

Despite this variability, no systematic trend with velocity is observed, supporting the use of a \emph{material-specific constant} for contact duration in safety governor design:

\begin{equation}
F_{est} = \frac{\Delta p}{\Delta t_{material}} = \frac{m \cdot v}{\Delta t_{material}}
\end{equation}

where $\Delta t_{material}$ is determined experimentally for each material.

\subsubsection{Critical Trade-off: Peak Force versus Rebound Energy}

Our analysis reveals a fundamental safety trade-off: configurations minimizing peak forces tend to maximize rebound energy. The bamboo design, while achieving the lowest peak force, exhibits the highest rebound (16.4 $\pm$ 2.9\%). 

\begin{figure}[h]
    \centering
    \includegraphics[width=\linewidth]{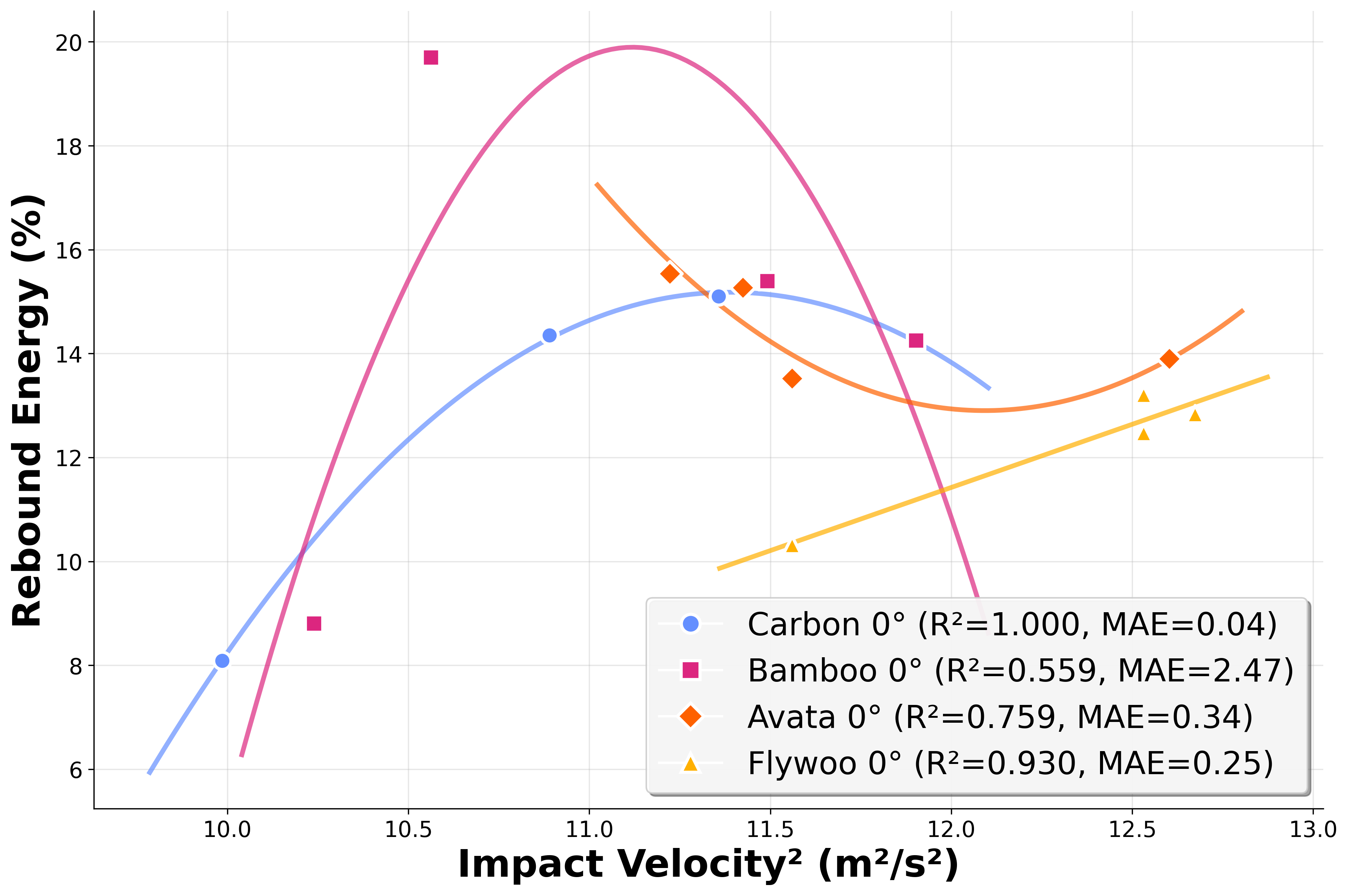}
    \caption{Rebound energy percentage versus configuration. Note inverse relationship with peak force magnitude.}
    \label{fig:rebound_0deg}
\end{figure}

\subsection{Angular Impact Effects and Real-World Relevance}

The 45° configuration reveals critical vulnerabilities in current safety assessments, which typically assume frontal impacts. Angular collisions increase peak forces by 27\% (105.6 $\rightarrow$ 133.9 N), approaching safety thresholds for impact with the neck and chest (ISO/TS 15066) even for compliant designs. This amplification results from:

\begin{table}[h]
\caption{Angular Impact Effects on Carbon Cognifly Performance}
\label{table_example}
\begin{center}
\begin{tabular}{|c||c|c|c|}
\hline
\rule{0pt}{3ex}\textbf{Parameter} & \textbf{0° Impact} & \textbf{45° Impact} & \textbf{Change}\rule[-1ex]{0pt}{0pt} \\
\hline
\rule{0pt}{2.5ex}$F_{max}$ (N) & 105.6 $\pm$ 7.0 & 133.9 $\pm$ 30.0 & +27\%\rule[-1ex]{0pt}{0pt} \\
\hline
\rule{0pt}{2.5ex}$\Delta t_J$ (ms) & 36.0 $\pm$ 3.3 & 38.3 $\pm$ 1.9 & +6\%\rule[-1ex]{0pt}{0pt} \\
\hline
\rule{0pt}{2.5ex}$J$ (N·s) & 1.144 $\pm$ 0.081 & 1.468 $\pm$ 0.204 & +28\%\rule[-1ex]{0pt}{0pt} \\
\hline
\rule{0pt}{2.5ex}$EC_r$ (\%) & 14.6 $\pm$ 0.5 & 11.2 $\pm$ 0.1 & -23\%\rule[-1ex]{0pt}{0pt} \\
\hline
\end{tabular}
\end{center}
\end{table}

The reduced rebound energy (-23\%) in angular impacts partially compensates for increased forces, suggesting that safety governors should incorporate orientation-dependent thresholds. The higher force variance ($\pm$30.0 N) indicates sensitivity to exact contact geometry, necessitating conservative safety margins for oblique collision scenarios.

\subsection{Design Factors for Safety Governor Implementation}
The dataset of our impact test results alongside the analysis scripts, the regressions output and the ROS2 Safety Governor nodes are available online\footnote{Link to be added after double-blind review.}.

\subsubsection{Impact Forces Within Operational Velocities}
Within the tested range, each configuration demonstrated consistent impact forces with coefficients of variation below 15\%. This consistency enables safety governors to employ fixed force thresholds for each drone configuration rather than velocity-dependent models within typical indoor operational speeds.

\subsubsection{Mass as Primary Safety Determinant}
Vehicle mass emerged as the dominant factor in impact severity. The 410g Avata generated 2.7× higher peak forces (230.4 N) compared to the 250g bamboo configuration (84.4 N), despite superior material properties. This relationship held across all configurations, confirming that mass reduction should be prioritized over material optimization for human-safe indoor UAVs.

\subsubsection{Quantified Benefits of Compliance Mechanisms}
Direct comparison between carbon configurations with and without compliance mechanisms revealed a 13\% force reduction: the carbon Cognifly with TPU joints (105.6 N) versus the rigid Flywoo design (120.8 N). This consistent reduction across the velocity range validates the integration of flexible elements in safety-critical designs.

\subsection{Safety Governor Validation}

\begin{figure}[h]
    \centering
    \includegraphics[width=1\linewidth]{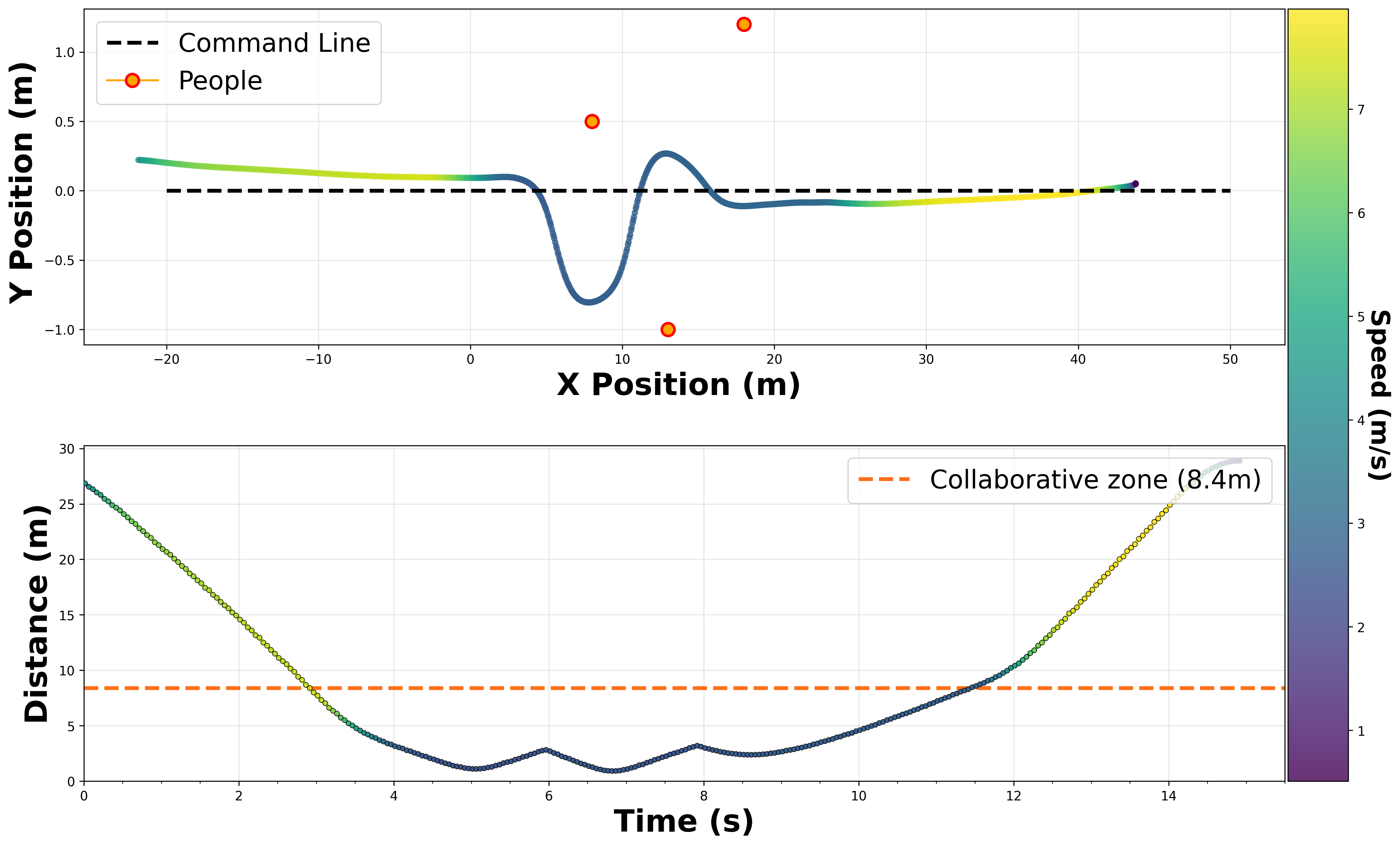}
    \caption{Safety Governor validation in Gazebo + ROS2 + PX4 SITL. Top image show the UAV trajectory across three people with velocity colormap. Bottom shows the distance to the nearest person with the same colormap.}
    \label{fig:safetygovernor}
\end{figure}

We validated the end-to-end integration of the impact characterization within a practical safety governor using a ROS~2 + Gazebo + PX4-SITL setup (Fig.~\ref{fig:safetygovernor}). The governed vehicle is the carbon \emph{Cognifly} (mass $m=0.25\,\text{kg}$), with worst-case deceleration $a=15\,\text{m/s}^2$, cruise $v_0=8\,\text{m/s}$, and a person-detection pipeline running at 10~Hz. The simulated behavior is a simple force field control with attraction points on opposite side of the field and three human considered as repulsion points. From Eq.~\ref{eq:iso-radius} (ISO/TS~15066 collaborative-space radius), the node computes $S\approx 8.4\,\text{m}$ around the person, within which the vehicle must converge to a safe speed. The force-aware cap is obtained from Eq.~\ref{eq:force-root} by evaluating the empirical $\hat e(v)=\sqrt{-3.601·Vi^2 + 82.013·Vi^2 - 451.789}$ (from Fig.\ref{fig:rebound_0deg}) and $\widehat{\Delta t}(v)=0.036s$ with a target average impact force $F^\star=140\,\text{N}$ at chest height, yielding $v_{\text{force}}\approx 3\,\text{m/s}$. When the simulated range $d(t)$ falls below $S$, the governor publishes \texttt{/cmd\_vel\_limited} with saturation shown in Fig.~\ref{fig:safetygovernor}.

While the present evaluation uses an outer-loop speed limiter, the same constraints can be embedded as state-dependent bounds in model predictive controllers. For example, Torrente et~al.~\cite{torrente2021} improve tracking by incorporating aerodynamic corrections within MPC, and Wang et~al.~\cite{wang2024} extend nonlinear MPC to suspended-payload flight with explicit handling of impact discontinuities. Our governor provides a complementary safety envelope: its distance- and impact-based limits can be injected as hard/soft constraints (or terminal penalties) in such MPC formulations, preserving feasibility and task performance while ensuring compliance with human-contact force bounds.




\section{CONCLUSION}
In this work we link benchtop impact characterization to a runtime safety feature for indoor MAVs operating near people. We built a compact, reproducible rig and protocol that measure force, impulse, and rebound on full airframes at indoor speeds, then fit simple regressions for restitution and contact duration. Combined with an ISO/TS 15066–style collaborative-space model, these fits drive a ROS 2 safety governor that enforces distance- and impact-based speed limits online, independent of the planner. We validated across multiple platforms and orientations and demonstrated deployment in PX4-SITL/Gazebo, showing conservative limits without sacrificing task feasibility. Released datasets, scripts, and the node form a practical toolchain for facility-specific policies grounded in measurement. Limitations include the current speed range and rotor-off tests; future work targets rotor-on impacts, online orientation compensation, MPC integration, and physical evaluations. Overall, the approach offers a measurement-backed path to safer, more certifiable indoor drone operations.

\section*{ACKNOWLEDGMENTS}
We thank Nicolas Beaudoin for supporting the Safety Governor simulation.



\bibliographystyle{IEEEtran}  
\bibliography{ICRA}


\end{document}